\documentclass[11pt]{article}
\pdfoutput=1
\usepackage{amsbsy,amsthm,amsmath,amssymb}
\usepackage{graphicx}
\usepackage[margin=1in]{geometry}
\usepackage{cleveref}

\usepackage{booktabs} 
\usepackage{multirow}

\usepackage{algorithm}
\usepackage{algpseudocode}

\usepackage{setspace}
\usepackage[american]{babel}
\usepackage[babel]{csquotes}
\usepackage{natbib}

\def\IEEEraisesectionheading{}
\def\IEEEPARstart{}
\def\color{}
\def\textcolor{}

\newtheorem{theorem}{Theorem}
\newtheorem{lemma}{Lemma}
\newtheorem{proposition}{Proposition}
\newtheorem{corollary}{Corollary}
\newtheorem{remark}{Remark}
\newtheorem{definition}{Definition}
\newtheorem{example}{Example}[section]

\DeclareMathOperator{\Cf}{\rm Cf}

\def\dist{\mathop{\rm dist}\nolimits}

\def\dom{\mathop{\bf dom}\nolimits}
\def\epi{\mathop{\bf epi}\nolimits}
\DeclareMathOperator*{\argmin}{arg\,min}


\renewcommand{\S}{Section~}

\allowdisplaybreaks

\title{On the Correctness of the Generalized Isotonic Recursive Partitioning Algorithm}

\author{
Joong-Ho Won\\
Department of Statistics\\
Seoul National University\and
Jihan Jung\\
SK Telecom\\
Seoul 04539, Korea
}

\date{}

\begin{document}
\maketitle

\begin{abstract}
This paper presents an in-depth analysis of the generalized isotonic recursive partitioning (GIRP) algorithm for fitting isotonic models under separable convex losses, proposed by Luss and Rosset [J. Comput. Graph. Statist., 23 (2014), pp. 192--201] for differentiable losses and extended by Painsky and Rosset [IEEE Trans. Pattern Anal. Mach. Intell., 38 (2016), pp. 308-321] for nondifferentiable losses. 
The GIRP algorithm poseses an attractive feature that in each step of the algorithm, the intermediate solution satisfies the isotonicity constraint.
The paper begins with an example showing that the GIRP algorithm as described in the literature may fail to produce an isotonic model, suggesting that the existence and uniqueness of the solution to the isotonic regression problem must be carefully addressed.
It proceeds with showing that, among possibly many solutions, 
there indeed exists a solution that can be found by recursive binary partitioning of the set of observed data. A small modification of the GIRP algorithm suffices to obtain a correct solution and
preserve the desired property that all the intermediate solutions are isotonic.
This proposed modification includes a proper choice of intermediate solutions and a simplification of the partitioning step from ternary to binary.

\end{abstract}

\onehalfspacing

\IEEEraisesectionheading{\section{Introduction}\label{sec:introduction}}
\IEEEPARstart{I}{sotonic} regression is an approach to fitting a function governing the relation between independent variables (covariates or predictors) and a dependent variable (response) under the assumption that the shape of the function preserves the order imposed on the covariates.
Such an assumption restricts the class of regression functions, which, in a simplest setting, leads to a collection of nondecreasing functions.
Regression modeling constrained by order is natrual in many applications including ranking, medicine, biology, and computer vision \citep{moon2010intervalrank,zheng2008query,schell1997reduced,obozinski2008consistent,weiss2019contrast}, where usual assumptions such as linearity are not so much applicable.
For example, the probability of a response to a treatment may increase with the treatment level \citep{robertson1988}, which is nonlinear due to saturation.

To fix the idea, let $X$ be a finite, non-empty, and partially ordered set of observed covariates.
We use $\preceq$ to denote the partial order.
Let $\mathbb{R}^X$ be the set of all real-valued functions from $X$ to $\mathbb{R}$.
Since $X$ is finite, $\mathbb{R}^X$ can be identified with the $|X|$-dimensional Euclidean space $\mathbb{R}^{|X|}$.
If we let the observed response for covariate $x \in X$ be $g(x)$, then $g\in\mathbb{R}^X$.
\begin{definition}[Isotonic function]
	Function $f\in\mathbb{R}^X$ is called \emph{isotonic} if 
for all $x, y \in X$, $x \preceq y$ implies $f(x) \leq f(y)$,
	where $\le$ has the usual meaning for real numbers.
\end{definition}
We denote the set of all isotonic functions in $\mathbb{R}^X$ by $\mathcal{I}_X$, which
is closed and convex \citep[p. 587]{spouge2003}.
Suppose a loss function
$\mathcal{J}_X\!:\!\mathbb{R}^{X}\!\times\!\mathbb{R}^{X} \!\to\! \mathbb{R}$
is given.
Then the isotonic regression problem is formally expressed as \begin{equation}\label{eqn:isotonic}
	\min\nolimits_{\hat{g}\in\mathcal{I}_X} \mathcal{J}_X(g, \hat{g}) 
.
	\end{equation}
It is not surprising that the most studied loss function is the squared error $\mathcal{J}_X(g, \hat{g})\!=\!\frac{1}{2}\sum_{x\in X}(g(x)\!-\!\hat{g}(x))^2$, i.e, the least squares regression. If the partial order is total, i.e., $x, y\!\in\!X$ implies either $x \preceq y$ or $y \preceq x$, then the pool-adjacent-violation algorithm (PAVA) computes the regression function in $O(|X|\log|X|)$ operations. For general partial orders, \citet{spouge2003} showed that it can be computed in up to $O(|X|^4)$ operations using a recursive partitioning algorithm based on the work of \citet{maxwell1985}.
The complexity is reduced, say, to $O(|X|^2)$ if the graph implied by the consraint set $\mathcal{I}_X$ is structured.
The essence of this algorithm is to fit the arithmetic mean of $g$ on the given set (initially $X$) and find a special kind of binary partition of it based on the fit.
In \citet{luss2012} it was noted that an early stopping of the recursive partitioning procedure always
yields a (suboptimal) isotonic function, an attactive feature that can be utilized to prevent overfitting.

Recently, \citet{luss2014,painsky2016} extended the isotonic recursive partitioning algorithm of \citet{maxwell1985,spouge2003,luss2012} to more general convex separable loss functions, i.e., $\mathcal{J}_X$ has the form $\mathcal{J}_X(g, \hat{g}) = \sum_{x\in X}\ell(g(x), \hat{g}(x))$, where the \emph{coordinate loss} $\ell(\cdot, \cdot): \mathbb{R}\times\mathbb{R} \to [0, \infty]$ is convex in the second argument for each value of the first argument. 
This extension, termed the \emph{generalized isotonic recursive partitioning} (GIRP) algorithm, 
fits a constant $b$ that minimizes the loss function restricted to the set $X$,
and partition $X$ into a triple $(L, M, U)$ that minimizes
\begin{equation}\label{eqn:painsky}
	\max\big(\sum\nolimits_{x\in U}\partial\ell(g(x), b) - \sum\nolimits_{x\in L}\partial\ell(g(x), b)\big)
\end{equation}
under the constraint that there exists a pair $(x, y)$ with $x \in L$ (resp. $M$) and $y \in M$ (resp. $U$) such that $x \preceq y$ but there is no such pair that $y \preceq x$.
If the minimum is nonnegative, then no partition is made.
Otherwise, the same procedure is applied to each of the triple recursively.
Here, $\partial\ell(g(x), b)$ denotes the subdifferential of $\ell(g(x), \cdot)$ at $b$, which is a closed interval, and the summation is in the Minkowski sense; see \Cref{sec:prelim} for details.
The GIRP algorithm has the same complexity as the least-squares case, and preserves the attractive feature that any intermediate solution preserves isotonicity. 
The coordinate loss functions $\ell$ considered in \citet{painsky2016} include the Huber loss
$$
	\ell_{\text{H},\delta}(a, b) = \begin{cases}
		\frac{1}{2}(a - b)^2, & |a - b| \le \delta, \\
		\delta|a - b| - \frac{1}{2}\delta^2, & \text{otherwise}
		,
	\end{cases}
$$
$\epsilon$-insensitive loss $\ell_{\epsilon}(a, b) = \max\{0, |a -b | - \epsilon\}$,
logistic loss $\ell_{\text{logit}}(a, b) = \log(1 + \exp(-ab))$, and hinge loss $\ell_{\text{hinge}}(a, b)=\max(0, 1 - ab)$. For the latter two losses, $a \in \{-1, +1\}$.

\begin{figure}[!t]
\centering
\includegraphics[width=2.8in]{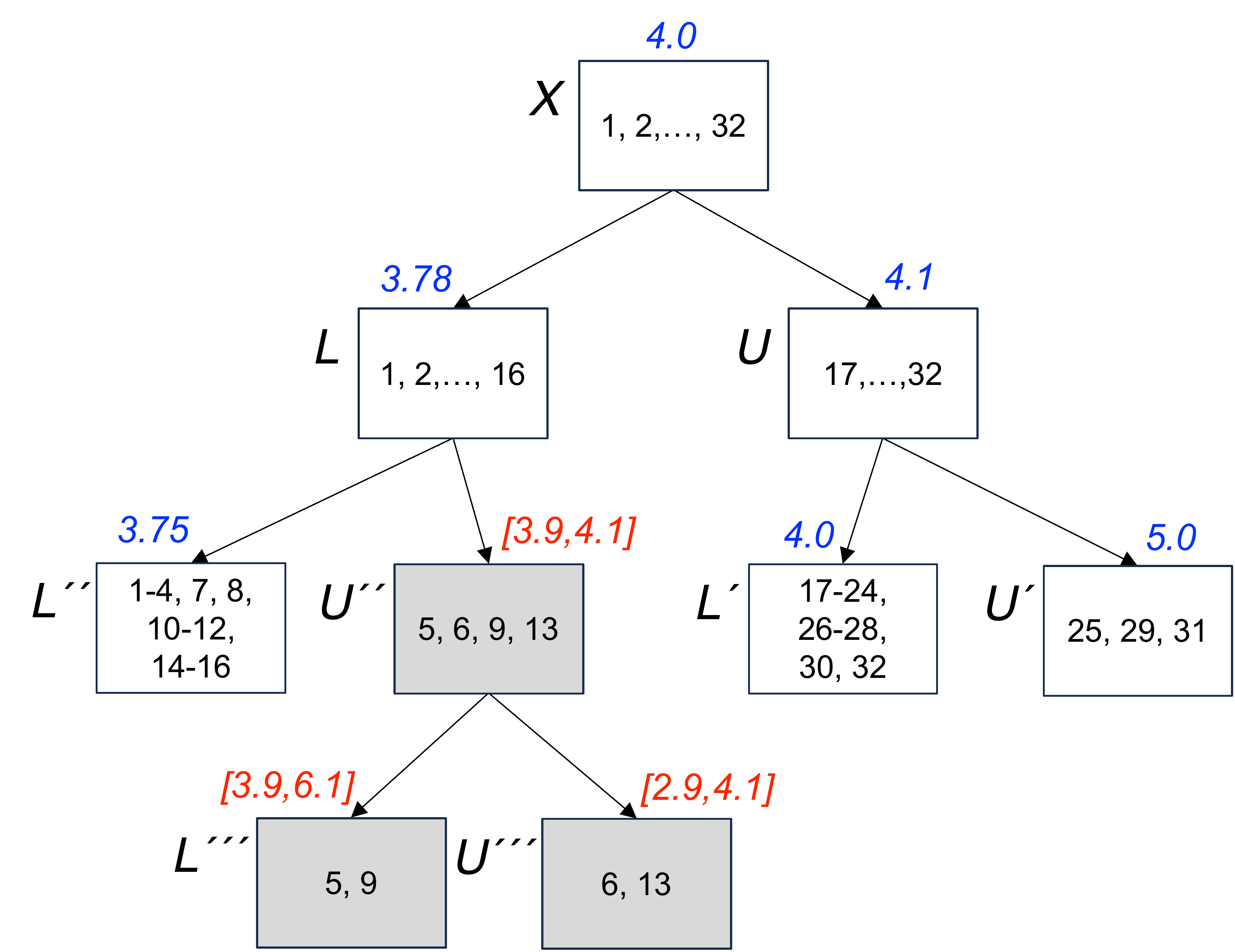}
\caption{Recursive partition tree from \Cref{ex:counter}. Shown inside each node (box) is the subset of $X$ representing the partition at its own level. On top of each box the fit to the set inside the box is presented in italic. Shaded nodes allow for nonunique fits.} \label{fig:counter}
\end{figure}

Unfortunately, however, there are combinations of data and coordinate loss functions that result in the failure of the algorithm.
Consider the following (counter-) example.
{\color{red}\begin{example}[Constraint violation]\label{ex:counter}
	Let
	$X = \{1, 2, \dotsc, 32\}$,
	\begin{align*}
		g = (&5, 4, 2, 7, 7, 5, 6, 6, 3, 4, 2, 4, 2, 2, 3, 1,  \\
	  	  &	4, 4, 4, 6, 4, 4, 4, 7, 8, 3, 4, 4, 5, 4, 3, 1)
		,
	\end{align*}
	and impose a partial order $\preceq$ represented by a directed graph $(X, E)$ with $E = \{(x, y): x \preceq y, ~x, y\in X\}$:
\begin{align*}
E = \{ &(1, 3), (5, 9), (2, 10), (4, 11), (7, 12), (8, 12), (10, 12), 
	\\
	&
		(4, 13), (6, 13), (9, 13), (8, 14), (10, 14), (12, 15), (12, 16), 
	\\
	&
	 (18, 22), (20, 22), (22, 25), (22, 26), (18, 27), (20, 27), 
	\\
	&
	 (23, 27), 
	 (21, 28), (24, 28), (18, 29),
	 (21, 29), (4, 30), 
	\\
	&
	 (6, 30), (9, 30), 
	 (25, 31), (26, 31),
	 (18, 32), (28, 32)\}
	.
\end{align*}
	If the Huber loss with $\delta=0.9$ is used, then the value $4.0$ is fitted to $X$, which is then partitioned into $L=\allowbreak \{1, 2, \dotsc, 16\}$, $M=\emptyset$, and $U=\{17, 18, \dotsc, 32\}$ by solving problem \eqref{eqn:painsky}. 
The fitted values are $3.78$ in $L$ and $4.1$ in $U$.
	Set $U$ is further split into $L'=\{17,\dotsc,24, 26, 27, 28, 30, 32\}$, $M'=\emptyset$, and $U'=\{25, 29, 31\}$.
	To the first set $4.0$ is fitted, and $5.0$ is fitted to the last set. 
	Set $L$ is further partitioned into $L''=\{1,2,3,4, 10, 11, 12, 14, 15, 16\}$, $M''=\emptyset$, and $U''=\{5, 6, 9, 13\}$, 
	The value $3.75$ is fitted to $L''$, and $4.1$ is fitted to $U''$.
	Then $U''$ is decomposed into $L'''=\{5, 9\}$, and $U'''=\{6, 13\}$ to which $6.1$ and $4.1$ is fitted respectively; $M'''=\emptyset$.
	This procedure yields
	\begin{align*}
	\hat{g} =& (3.75, 3.75, 3.75, 3.75, 6.1, 4.1, 3.75, 3.75, 6.1, 3.75, 3.75, 
	\\
	&~ 3.75, 4.1, 3.75, 3.75, 3.75, 4.0, 4.0, 4.0, 4.0, 4.0, 4.0, 4.0, 
	\\
	&~ 4.0, 5.0, 4.0, 4.0, 4.0, 5.0, 4.0, 5.0, 4.0 )
	.
	\end{align*}
	However, $\hat{g}(6) = 4.1$, $\hat{g}(9) = 6.1$, and $\hat{g}(30) = 4.0$, violating the order constraints $\hat{g}(6) \le \hat{g}(30)$ and $\hat{g}(9) \le \hat{g}(30)$.
	Furthermore, if the recursive partitioning procedure is terminated early and $U''$ is not split, then we have $\hat{g}(5)=\hat{g}(6)=\hat{g}(9)=\hat{g}(13)=4.1$ and the cited order constraints are again violated.
	The detailed computations are worked out in \Cref{sec:examples}.
\end{example}
\begin{example}[Frequency of violation]\label{ex:numeric}
For more numerical evidences, a simulation study is conducted.
Following the design in \cite[\S3.2]{luss2014}, two-dimensional covariates were generated randomly from a uniform distribution and isotonic constraints were derived. Responses were drawn from Poisson and polynomial regression models.
We varied the $\delta$ in $\ell_{\text{H},\delta}$ and the sample size $n$ and counted the number of violations of the isotonic constraints in 100 repeated fits for each $(\delta, n)$ pair.
Four models were considered:
\begin{itemize}
	\item Model 1: $y_i \sim \text{Poisson}(\prod_{j=1}^d x_{ij}^{1/2})$, $x_{ij} \sim \text{unif}[0, 10]$,
	\item Model 2: $y_i \sim \text{Poisson}(\prod_{j=1}^d x_{ij}^{2})$, $x_{ij} \sim \text{unif}[5, 10]$,
	\item Model 3: $y_i \sim \prod_{j=1}^d x_{ij} + N(0, 4)$, $x_{ij} \sim \text{unif}[0, 3]$,
	\item Model 4: $y_i \sim \prod_{j=1}^d x_{ij}^2 + N(0, 9)$, $x_{ij} \sim \text{unif}[0, 5]$,
\end{itemize}
for $i=1,\dotsc,n$.
\Cref{tbl:violation} illustrates that as $\delta$ decreases, the number of violations increases. Furthermore, it is evident that violations tend to increase with larger sample sizes.
\end{example}

To see the reason for these violations,
let us focus on the set $U''=\{5, 6, 9, 13\}$ in \Cref{ex:counter}.
Since 
$g$ takes values $7, 5, 3, 2$ on this set,
the GIRP algorithm fits to this set a value that
minimizes \begin{align*}
	f_{U''}(b) 
	&= \ell_{\text{H},.9}(7, b) 
	+ \ell_{\text{H},.9}(5, b) 
+ \ell_{\text{H},.9}(3, b) 
	+ \ell_{\text{H},.9}(2, b) 
	.
\end{align*}
This convex function is piecewise quadratic, and takes its minimum on the interval $[3.9, 4.1]$. Thus any $b\in[3.9, 4.1]$ can be fitted. 
Likewise, for $L'''$ and $U'''$ any value in $[3.9, 6.1]$ and $[2.9, 4.1]$ can be fitted, respectively.
The partition tree induced by the GIRP algorithm is shown in \Cref{fig:counter}. 
Although the fits can be chosen not to violate the isotonic constraints (e.g., $3.95$ to $U''$, $3.9$ to $L'''$, $4.0$ to $U'''$), they cannot be determined without traversing the opposite subtree. For example, the fit to $U''$ must not be greater than $4.0$, the smallest fit in the subtree rooted at $U$.
Since recursive partitioning cannot backtrack beyond the input partition, such a scenario is not feasible.
Simple rules such as taking the smallest, largest, or middle values will not resolve the problem either.  \Cref{ex:counter} chose the largest. The other two choices all result in a violation of isotonicity as well.}

This counterexample suggests that existence and uniqueness of the fit plays an important role in the correctness of the GIRP algorithm. 
In the least-squares setting \citep{maxwell1985,spouge2003,luss2012}, 
the loss function is differentiable and
the fit (arithmetic mean) is always unique. 
In general, however,
complex interactions of coordinate loss functions may result in 
nonuniqueness of the fit
even if each function
is uniquely minimized.
Beyond Huber, both the $\epsilon$-insensitive and hinge losses are nondifferentiable and possess infinitely many minimizers. To worsen the situation, the logistic loss does not attain a minimum. Unfortunately, the analyses of \citet{luss2014,painsky2016} assume implicitly that the fits are unique.

\textcolor{red}{These unpleasant scenarios raise two fundamental questions: 1) What are the conditions under which a solution to problem \eqref{eqn:isotonic} exists, and is it unique?
2) Given the existence of potentially multiple solutions, is there a solution that can be identified using a recursive partitioning algorithm?
The goal of the present paper is to answer these questions. 
After answering them affirmatively, 
we proceed to show that a simple modification of the GIRP algorithm suffices to find such a solution, with the desired properties that \emph{no backtracking is required} and \emph{any intermediate solution is isotonic}.}

Our analysis is inspired by the approach of \citet{spouge2003}. Compared with \citet{luss2012,luss2014,painsky2016} that are based on the Karush-Kuhn-Tucker conditions, our approach is based on the first principle, hence provides a clearer picture to the problem. Also, the notions of ``groups'' and ``blocks'' are avoided and the partitioning algorithm is presented solely in terms of upper and lower sets, standard notions in order theory \citep{robertson1988}. The consequence is a unifying treatment to the graph cut problem for partitioning under both differentiable and non-differentiable losses, unlike the differing strategies of \citet{luss2014} (binary partitioning) and \citet{painsky2016} (ternary partitioning).

After developing a necessary background in \Cref{sec:prelim} to answer the first question,
\Cref{sec:decomposition} shows that a solution to problem \eqref{eqn:isotonic} can be found by recursive partitioning even in the absence of uniqueness, answering the second.
\Cref{sec:girp} discusses the
modification of the GIRP algorithm that correctly computes the solution.
Illustrative examples are given in \Cref{sec:examples}, 
after which the paper is concluded.

\begin{table}[h]
\caption{Frequency of violation of isotonic constraints}\label{tbl:violation}
\scriptsize
\centering
\begin{tabular}{lrrrrrr}
\toprule
\multirow{2}{*}[-2pt]{model} & \multirow{2}{*}[-2pt]{sample size} & \multicolumn{5}{c}{$\delta$} \\
\cmidrule{3-7}
& & $0.1$ & $0.5$ & $0.9$ & $3.0$ & $5.0$ \\
\midrule
\multirow[t]{4}{*}{Model 1} 
 &    50 &  67 &  69 &  55 &   1 &   0 \\
 &   100 &  96 &  94 &  87 &   6 &   0 \\
 &  1000 & 100 & 100 & 100 &  45 &   4 \\
 &  5000 & 100 & 100 & 100 &  89 &  12 \\
\multirow[t]{4}{*}{Model 2} 
 &    50 &  54 &  54 &  53 &  34 &  18 \\
 &   100 &  95 &  94 &  93 &  83 &  57 \\
 &  1000 & 100 & 100 & 100 & 100 & 100 \\
 &  5000 & 100 & 100 & 100 & 100 & 100 \\
\multirow[t]{4}{*}{Model 3} 
 &    50 &  96 &  91 &  89 &  31 &   4 \\
 &   100 &  99 &  98 &  95 &  65 &  16 \\
 &  1000 & 100 & 100 & 100 &  98 &  63 \\
 &  5000 & 100 & 100 & 100 & 100 &  97 \\
\multirow[t]{4}{*}{Model 4} 
 &    50 &  94 &  88 &  87 &  72 &  47 \\
 &   100 & 100 & 100 &  99 &  89 &  71 \\
 &  1000 & 100 & 100 & 100 & 100 & 100 \\
 &  5000 & 100 & 100 & 100 & 100 & 100 \\
\bottomrule
\end{tabular}
\end{table} 
 \section{Preliminaries}\label{sec:prelim}
We begin with basic notation. Any convex function $f$ on $\mathbb{R}^n$ considered in this paper is extended-real-valued, i.e., its range can be in $\mathbb{R}\cup\{\infty\}$. Its essential domain refers to $\dom f = \{x\in\mathbb{R}^n: f(x) < \infty\}$. 
Function $f$ is proper if $\dom f \neq \emptyset$.
The epigraph of $f$ is $\epi f = \{(x, t)\in\mathbb{R}^n\times\mathbb{R}: t \ge f(x) \}$.
The subdifferential of $f$ at $x\in\mathbb{R}^n$ is the set of subgradients $\partial f(x) = \{s: f(y) \ge f(x) + \langle s, y - x \rangle \}$; it is compact if $x \in \dom f$.
If $f(x) = \alpha_1 f_1(x) + \alpha_2 f_2(x)$ for convex $f_1$ and $f_2$ with $\alpha_1, \alpha_2 \ge 0$, then $\partial f(x) = \alpha_1 \partial f_1(x) + \alpha_2 \partial f_2(x)$. 
Here the summation is in the Minkowski sense, i.e.,
$A + B \triangleq \{a + b: a \in A, ~b \in B \}$
and
$kA \triangleq \{ka: a \in A\}$,
	~ $k \in \mathbb{R}$.
We write $-A = (-1)A$ and $A - B = A + (-B)$.
The set of minimizers of $f$ on the constraint set $C\subset\mathbb{R}^n$ is denoted by $\argmin_{x\in C} f(x)$. When this set is a singleton, we abuse the notation and write, e.g., $x^{\star} = \argmin_{x\in C}f(x)$.

For a real-valued function $g \in \mathbb{R}^X$ on $X$, its restriction to set $S \subset X$ is denoted by $g|_S$. 
If $(L, U)$ is a partition of $X$, then $g \!=\! g|_L \cup g|_U$ means that $g(x) \!=\! g|_L(x)$ for all $x \in L$ and $g(x) \!=\! g|_U(x)$ for all $x \in U$.
Let $\chi_S \in \mathbb{R}^X$ be the characteristic function such that $\chi_S(x) = 1$ if $x \in S$ and $\chi_S(x)=0$ if $x \not\in S$.
A constant function taking the value $k\in\mathbb{R}$ in its domain is denoted by $k\mathbf{1}$. 
If two functions $g_1, g_2 \in \mathbb{R}^S$ are such that $g_1(x) \!\ge\! g_2(x)$ for all $x\in S$, then we write $g_1 \ge g_2$; $g_1 \le g_2$ and $g_1 = g_2$ are analogous. 

We consider separable loss functions in problem \eqref{eqn:isotonic}, i.e., $\mathcal{J}_X$ has the form
$\mathcal{J}_X(g, \hat{g}) = \sum_{x\in X}\ell(g(x),\allowbreak \hat{g}(x))$, 
where the coordinate loss $\ell(\cdot, \cdot): \mathbb{R}\times\mathbb{R} \to [0, \infty]$ is convex in the second argument for each first argument. 
We further assume that function $f(y) = \ell(g(x), y)$ is lower semicontinuous (i.e., $\epi f$ is closed)
and $\dom f$ 
is an interval (open endpoint of $\pm\infty$ is allowed) but not a singleton.
Hence $f$ is a closed, convex, and proper function.
For any $S \subset X$, we write 
$\mathcal{J}_S(g, \hat{g}) = \sum_{x\in S}\ell(g(x), \hat{g}(x))$.

Since $\mathbb{R}^X$ is identified with an Euclidean space $\mathbb{R}^{|X|}$ and the set of isotonic functions $\mathcal{I}_X \subset \mathbb{R}^X$ is closed and convex, the following is a standard result from optimization theory.
\begin{proposition}[Existence and uniqueness of solution]\label{prop:existence}
	If $\mathcal{I}_X$ is nonempty and bounded, then problem \eqref{eqn:isotonic} has a solution.
	If $f = \ell(g(x), \cdot)$ is coercive, i.e., $\lim_{|y|\to\infty} f(y) = \infty$, for each $x$, then problem \eqref{eqn:isotonic} has a solution whether or not $\mathcal{I}_X$ is bounded.
	If $f$ is strictly convex and coercive, then the solution is unique.
\end{proposition}
For least-squares isotonic regression, $f = \ell(g(x),\cdot)=\frac{1}{2}(g(x) - \cdot)^2$ is strongly convex, hence strictly convex and coercive. Thus the solution is unique.
The Huber and $\epsilon$-insensitive losses are coercive but not strictly convex, so uniqueness is not guaranteed.
For the hinge and logistic losses, $f$ is not coercive. So if $\mathcal{I}_X$ is unbounded in such a way that the direction $d \in \mathbb{R}^X$ with which $\lim_{t\to\infty}\mathcal{J}_X(g, \hat{g}_0 + t d) = 0$ is feasible for some $\hat{g}_0 \in \mathcal{I}_X$ (which occurs when the data is perfectly separable),
then the solution may not exist. A simple example is when $X=\{1, 2\}$, $g(1) = 1$, $g(2) = -1$, and $\mathcal{I}_X = \{\hat{g}: \hat{g}(1) \ge \hat{g}(2)\}$.
With the logistic loss, indeed no solution exists.
For the hinge loss, however, since there is a \emph{bounded} minimizing sequence $\{y_k\}$ of $f$, i.e., $\lim_{k\to\infty}f(y_k) = \inf_{y\in\mathbb{R}}f(y) = 0$, the minimum is attained even though \Cref{prop:existence} does not guarantee it;
the minimizer is not unique though.

In the sequel, we assume that a solution to problem \eqref{eqn:isotonic} exists, albeit possibly not unique.
In particular, perfectly separable cases with the logistic loss are ruled out.

For a function $f \in \mathbb{R}^X$, let us denote by $[f < a]$ the set $\{x\in X: f(x) < a\}$. Sets $[f \le a]$, $[f > a]$, $[f \ge a]$, $[f = a]$, and $[f \neq a]$ are defined analogously. 
The following lemma extends Theorem 1.3.5 of \citet{robertson1988} for separable convex losses.

\begin{definition}[Constant fit]
	Given $X$, $g$, and $\ell$ as described and any nonempty set $S \subset X$, let a constant fit of $g$ on $S$ be
	an element of the set
	$$
\Cf_g(S) \triangleq \argmin_{y\in\mathbb{R}}\sum\nolimits_{x\in S}\ell(g(x), y) = \argmin_{y\in\mathbb{R}} \mathcal{J}_S(g, y\mathbf{1})
		.
	$$
\end{definition}

\begin{lemma}\label{lem:min}
	Suppose $g^{\star}$ minimizes $\mathcal{J}_X(g, \cdot)$ over $\mathcal{I}_X$.
	If $c$ is any real number and if the set $[g^{\star} = c]$ is nonempty then 
$c \in \Cf_g([g^{\star}=c])$.
\end{lemma}
\begin{proof}
	Clearly,
	\begin{align*}
		\mathcal{J}_X(g, g^{\star}) &= \sum\nolimits_{x\in [g^{\star}\!\neq\! c]}\!\ell(g(x), g^{\star}(x))
		+ \sum\nolimits_{x\in [g^{\star}\!=\! c]}\!\ell(g(x), c)
		.
	\end{align*}
	Let $M = \argmin_{y\in\mathbb{R}} \sum_{x\in[g^{\star}=c]}\ell(g(x), y)$.
	If $M=\emptyset$, then $f(y) = \sum_{x\in[g^{\star}=c]}\ell(g(x), y)$ in strictly monotone on $\mathbb{R}$ and
	there exists $\tilde{c} \neq c$ such that $f(\tilde{c}) < f(c)$. Such $\tilde{c}$ can be chosen sufficiently close to $c$ so that function $\tilde{g} \in \mathbb{R}^{X}$ that is equal to $g^{\star}$ on $[g^{\star}\neq c]$ and equal to $\tilde{c}$ on $[g^{\star}=c]$ is isotonic, hence in $\mathcal{I}_X$. It follows that $\mathcal{J}_X(g, \tilde{g}) < \mathcal{J}_X(g, g^{\star})$, contradicting the hypothesis that $g^{\star}$ minimizes $\mathcal{J}_X(g, \cdot)$.

	Thus $M= \Cf_g([g^{\star}\!=\!c])$ is nonempty. It takes the form of an interval since $\ell(g(x), \cdot)$ is convex. If $c \notin M$, then $\dist(c, M) \triangleq \inf_{m\in M}|c - m| > 0$. 
Choose $\tilde{c} \notin M$ sufficiently close to $c$ such that $\dist(c, M) > \dist(\tilde{c}, M) > 0$ and $\tilde{g}\in\mathbb{R}^X$ defined as above is isotonic. Once again it follows that $\mathcal{J}_X(g, \tilde{g}) \!<\! \mathcal{J}_X(g, g^{\star})$, a contradiction.
	Hence $c \in M$. \end{proof}

The following notion, taken from \citet{robertson1988}, plays an important role in the sequel.
\begin{definition}[Upper set]
	A subset $U$ of $X$ is called an \emph{upper set} if $x\in U$, $y\in X$, and $x\preceq y$ implies $y\in U$. That is, an upper set contains all elements ``greater than or equal to'' any of its element. A \emph{lower set} is defined similarly in a dual fashion. \end{definition}
A set $U$ is an upper set if and only if $L=U^c=X\setminus U$ is a lower set. 
We denote the collection of all upper sets of $X$ by $\mathcal{U}_X$ and that of all lower sets by $\mathcal{L}_X$.
Note that $\emptyset$ and $X$ belong to $\mathcal{U}_X$ and $\mathcal{L}_X$ simultaneously.
Collections $\mathcal{U}_X$, $\mathcal{L}_X$ and the set $\mathcal{I}_X$ of all isotonic functions on $X$ are related as follows.
\begin{proposition}[Theorem 1.4.1 of \cite{robertson1988}]\label{prop:upperset}
	A function $f\in\mathbb{R}^X$ is isotonic if and only if one of the following conditions is satisfied:
	\begin{enumerate}
	\item $[f < a]\in\mathcal{L}_X$ for any $a \in \mathbb{R}$;
	\item $[f \le a]\in\mathcal{L}_X$ for any $a \in \mathbb{R}$;
	\item $[f > a]\in\mathcal{U}_X$ for any $a \in \mathbb{R}$; or
	\item $[f \ge a]\in\mathcal{U}_X$ for any $a \in \mathbb{R}$.
	\end{enumerate}
\end{proposition}

 \section{Recursive binary partitioning}\label{sec:decomposition}

\textcolor{red}{The goal of this section is to show that a solution to problem \eqref{eqn:isotonic} can be found by solving reduced versions of \eqref{eqn:isotonic} on special lower and upper sets in $X$ independently, which together forms a \emph{projection pair} (\Cref{thm:decomp}). This suggests a recursive binary partitioning procedure.
The problem of finding a projection pair is converted to that of finding a \emph{maximal upper set} (\Cref{thm:3_2}).
Thus a procedure that seeks a maximal upper set in the given partition recursively 
with a proper terminating case (\Cref{thm:constantfit}) should solve \eqref{eqn:isotonic}.}

\textcolor{red}{Compared with \citet{spouge2003} for least squares, the challenge here is to deal with nonuniqueness. To do this, we need to generalize the notion of projection pair \citep[Def. 3.1]{spouge2003}:}
\begin{definition}[Complementary pair] A pair $(L, U)$ such that $U$ is an upper set in $X$ and $L=X \setminus U$ (guaranteed to be a lower set) is called a \emph{complementary pair} of $X$. The pair is \emph{trivial} if either $U$ or $L$ is empty.
\end{definition}

\begin{definition}[Projection pair]\label{def:projectionpair}
A nontrivial complementary pair in $X$ is a \emph{projection pair} for $g \in \mathbb{R}^X$ with separation constant $c \in \mathbb{R}$ if there exist 
$(g|_L)^{\star}\!\in\!\argmin_{\hat{g}\in\mathcal{I}_L}\!L_L(g|_L, \hat{g})$,
$(g|_U)^{\star}\!\in\!\argmin_{\hat{g}\in\mathcal{I}_U}\!L_U(g|_U, \hat{g})$,
and
$c$ such that 
$(g|_L)^{\star} \le c\mathbf{1} \le (g|_U)^{\star}$.
\end{definition}
\textcolor{red}{\Cref{thm:decomp} extends \citet[Thm. 3.1]{spouge2003} by showing that \eqref{eqn:isotonic} is indeed solved by solving the isotonic regression problems independently for the above notion of projection pair.}
\begin{theorem}\label{thm:decomp}
	If $(L, U)$ is a projection pair for $g$ with 
	separation constant $c$,
	then $(g|_L)^{\star} \cup (g|_U)^{\star}$ solves problem \eqref{eqn:isotonic}.
\end{theorem}
\begin{proof}
	Let $g_{LU}^{\star}(x) = (g|_L)^{\star}(x)$ if $x \in L$ and $g_{LU}^{\star}(x) = (g|_U)^{\star}(x)$ if $x \in U$.
	Suppose $g^*$ minimizes $\mathcal{J}_X(g, \cdot)$ over $\mathcal{I}_X$.
	Then, due to separability,
	\begin{align*}
		\mathcal{J}_X(g, g^*) &= \mathcal{J}_L(g, g^*) + \mathcal{J}_U(g, g^*) \\
					&\ge \mathcal{J}_L(g, (g|_L)^{\star}) + \mathcal{J}_U(g, (g|_U)^{\star}) \\
					&= \mathcal{J}_L(g, g_{LU}^{\star}) + \mathcal{J}_U(g, g_{LU}^{\star}) = \mathcal{J}_X(g, g_{LU}^{\star})
		.
	\end{align*}
	Since $(g|_L)^{\star} \le c\mathbf{1} \le (g|_U)^{\star}$, $g_{LU}^{\star}$ is isotonic. Hence $g_{LU}^{\star}$ minimizes $\mathcal{J}_X(g, \cdot)$ over $\mathcal{I}_X$.
\end{proof}
\begin{remark}
	\color{red}
	Unlike \citet[Thm. 3.1]{spouge2003},
	it is not necessarily true that $g^* = g_{LU}^{\star}$ since uniqueness is not guaranteed. 
\end{remark}

\textcolor{red}{A natural next step is to find a projection pair.
For the least squares problem, \citet{spouge2003,maxwell1985} reduce this problem to that of finding a maximal upper set that maximizes the negated derivative of the partial sum of coordinate losses over all upper sets.}
We generalize this approach to deal with subdifferentials, which are set-valued. 
Let $b$ be any real number and define a set-valued function on $X$ to $2^{\mathbb{R}}$ by
$$
\Sigma_b(S) = -\sum\nolimits_{x\in S} \partial\ell(g(x), b)
	,
	\quad
	S \subset X
	.
$$
Since $\partial\ell(g(x), b)$ is a closed and bounded interval, computing $\Sigma_b(S)$ involves interval arithmetic. For example, if $I_1$ and $I_2$ are two closed (not necessarily bounded) intervals, then $\min\{I_1 + I_2\} = \min I_1 + \min I_2$.\footnote{If $I=(-\infty, a]$, then take $\min I = -\infty$.}
\textcolor{red}{Now let us define our notions of maximal upper set and its dual.}
\begin{definition}[Maximal upper set, minimal lower set]\label{def:maximalupper}
A \emph{maximal upper set $U$ with respect to $b$} is an upper set such that $\min\Sigma_b(U)=\max_{U'\in\mathcal{U}_X} \min\Sigma_b(U')$. A \emph{minimal lower set} is defined in a dual fashion.
\end{definition}

\textcolor{red}{The following theorem, which generalizes \citet[Thm. 3.2]{spouge2003}, asserts that a maximal upper set (or minimal lower set) as defined above determines a projection pair.}
\begin{theorem}\label{thm:3_2}
	For a nontrivial complementary pair $(L, U)$, if $U$ is a maximal upper set with respect to $b$, then
	$(L, U)$ is a projection pair for $g$
	with the separation constant $b$.
	Alternatively, if $L$ is a minimal lower set with respect to $b$, the same conclusion follows.
\end{theorem}
\begin{proof}
	We focus on the case that $U$ is a maximal upper set. A dual argument applies if $L$ is a minimal lower set.
	Pick a $(g|_U)^{\star}$ that minimizes $\mathcal{J}_U(g, \cdot)$ over $\mathcal{I}_U$.
	Set $\underline{u} = \min_{x\in U}(g|_U)^{\star}(x)$.
	Note that $U$ can be written as the disjoint union
$[(g|_U)^{\star}=\underline{u}]\cup[(g|_U)^{\star}>\underline{u}]$.
Obviously, 
	$$
		\Sigma_b(U) = \Sigma_b([(g|_U)^{\star}=\underline{u}]) + \Sigma_b([(g|_U)^{\star}>\underline{u}])
		.
	$$
	A simple interval arithmetic entails
	$$
		\min\Sigma_b(U) = \min\Sigma_b([(g|_U)^{\star}=\underline{u}]) + \min\Sigma_b([(g|_U)^{\star}>\underline{u}])
		.
	$$
	Since $[(g|_U)^{\star}>\underline{u}]$ is an upper set in $X$ (see \Cref{prop:upperset}), 
the maximality of $U$ entails
	$$
		\min\Sigma_b([(g|_U)^{\star}>\underline{u}]) \le \min\Sigma_b(U)
		.
	$$
	which leads to
	\begin{equation}\label{eqn:sigma}
	\begin{split}
		0 &\le \min\Sigma_b([(g|_U\!)^{\star}\!=\!\underline{u}])
		\!=\! -\!\max\!\sum\nolimits_{[(g|_U\!)^{\star}\!=\!\underline{u}]}\!\partial\ell(g(x),\! b) 
		\\
		&= -\max\partial \mathcal{J}_{[(g|_U)^{\star}=\underline{u}]}(g, b\mathbf{1})
		.
	\end{split}
	\end{equation}
	This means that the univariate convex function $f(y) = \mathcal{J}_{[(g|_U)^{\star}=\underline{u}]}(g, y\mathbf{1})$ is nonincreasing at $y=b$.

	Lemma \ref{lem:min} implies that $\underline{u}$ minimizes $f(y)$. If $\min\Sigma_b([(g|_U)^{\star}=\underline{u}]) > 0$, then $f(y)$ is strictly decreasing at $b$. It follows from convexity that 
$b \le \underline{u} \le (g|_U)^{\star}(x)$, $\forall x \in U$.

	If $\min\Sigma_b([(g|_U)^{\star}=\underline{u}]) = 0$, then $0 \in \Sigma_b([(g|_U)^{\star}=\underline{u}])=\partial \mathcal{J}_{[(g|_U)^{\star}=\underline{u}]}(g, b\mathbf{1})$. Hence $b$ also minimizes $f(y)$ but no order can be asserted between $b$ and $\underline{u}$ unless the minimizer of $f(y)$ is unique (this occurs, e.g, if $\ell(g(x), \cdot)$ is strictly convex for each $x$).
If $\underline{u} < b$, then any value in $[\underline{u}, b]$ is a minimizer of $f(y)$.
	Now note that 
		$$\min\Sigma_b([(g|_U)^{\star}>\underline{u}]) = \min\Sigma_b(U)$$
	and the set $[(g|_U)^{\star}>\underline{u}]$ is also maximal. 
	Thus we can replace $U$ with a strictly smaller set $U'=[(g|_U)^{\star}>\underline{u}]$ and then repeat the preceding analysis. 
	This procedure is terminated when $\min\Sigma_b([(g|_{U'})^{\star}\!=\!\underline{u}'])\!>\!0$ or $U'=\emptyset$, where $\underline{u}' = \min_{x\in U'}(g|_{U'})^{\star}(x)$ minimizes $f(y) = \mathcal{J}_{[(g|_{U'})^{\star}>\underline{u}']}(g, y\mathbf{1})$. In the former case $f(y)$ is strictly decreasing at $y=b$ hence $b < \underline{u}'$. In the latter case $(g|_{U'})^{\star}$ is irrelevant and we can set $\underline{u}'=b$.
	Thus for the final $U'$, we have $b \le \underline{u}' \le (g|_{U'})^{\star}(x)$ for all $x\in U'$.
	Now $U\setminus U'$ is not empty, and by construction any distinct value $v$ of $(g|_U)^{\star}(x)$ on this set lies in $[\underline{u}, b]$ and minimizes $f_v(y) \triangleq \mathcal{J}_{[(g|_U)^{\star}=v]}(g, y\mathbf{1})$. At the same time, $b$ also minimizes $f_v$.
	There are only a finite number of such $v$'s in $[\underline{u}, b]$, say, $v_1, \dotsc, v_k$. Since $(g|_U)^{\star}$ minimizes $\mathcal{J}_U(g, \cdot)$,
\begin{align*}
	&\mathcal{J}_U(g, \hat{g}) = \sum\nolimits_{x\in[(g|_U\!)^{\star}\!=\!\underline{u}]}\ell(g(x), \hat{g}(x))
	\\
	&
	\, +\!\sum\nolimits_{j\!=\!1}^k\sum\nolimits_{x\in[(g|_U\!)^{\star}\!=\!v_j]}\!\ell(g(x), \hat{g}(x))
	\!+\! \sum\nolimits_{x\in[(g|_U\!)^{\star}\!>\!\underline{u}]}\!\ell(g(x), \hat{g}(x))
	\\
	&\ge\sum\nolimits_{x\in[(g|_U\!)^{\star}\!=\!\underline{u}]}\!\ell(g(x), \underline{u})
	+ \sum\nolimits_{j\!=\!1}^k\!\sum\nolimits_{x\in[(g|_U\!)^{\star}\!=\!v_j]}\!\ell(g(x), v_k)
	\\
	&\quad 
	+ \sum\nolimits_{x\in[(g|_U)^{\star}>\underline{u}]}\ell\big(g(x), (g|_U)^{\star}(x)\big)
	\\
	&\stackrel{(*)}{=}\sum\nolimits_{x\in[(g|_U\!)^{\star}\!=\!\underline{u}]}\!\ell(g(x), b)
	+\sum\nolimits_{j!=\!1}^k\!\sum\nolimits_{x\in[(g|_U\!)^{\star}\!=\!v_j]}\!\ell(g(x), b)
	\\
	&\quad + \sum\nolimits_{x\in[(g|_U)^{\star}>\underline{u}]}\ell\big(g(x), (g|_U)^{\star}(x)\big)
	= \min_{\hat{g}\in \mathcal{I}_{U}}\mathcal{J}_U(g, \hat{g})
	.
\end{align*}
Equality (*) means that $(g|_U)^{\dagger} = \max\{b\mathbf{1}, (g|_U)^{\star}\}$
	also minimizes $\mathcal{J}_U(g, \cdot)$. Then we have
$$
		b \le \underline{u}' \le (g|_U)^{\dagger}(x), 
		\quad \forall x \in X
		.
$$

	A parallel argument shows that for $(g|_L)^{\dagger} = \min\{b\mathbf{1}, (g|_L)^{\star}\}$, we have
$(g|_U)^{\dagger}(x) \le \bar{l}' \le b,\forall x \in X$,
where $\bar{l}' = \min_{x\in[(g|_L)^{\star} < b]}(g|_L)^{\star}(x)$.
Then $(g|_L)^{\dagger} \leq b\mathbf{1} \leq (g|_U)^{\dagger}$ and
it follows that $(L, U)$ is a projection pair.
\end{proof}
\begin{remark}\label{rem:spouge}
	\color{red}
Unlike \citet[Thm. 3.2]{spouge2003}, that $U$ is a maximal upper set does not necessarily mean that $L=X\setminus U$ is a \emph{minimal} lower set, unless $\Sigma_b(S)$ is a singleton for every possible partition $S \subset X$.
\end{remark}
\begin{remark}\label{rem:rosset}
	The claim that GIRP provides a solution to problem \eqref{eqn:isotonic} in \citet{luss2014,painsky2016} is essentially based on the assumption that $b \le \underline{u}$, which cannot be guaranteed unless the $f(y)$ in the proof of \Cref{thm:3_2} is uniquely minimized.
   	(In inequality (40) of \citet{painsky2016} and inequality (1) of the Appendix of \citet{luss2014}, the possibility of the subgradient $g$ being zero cannot be ruled out. If $g=0$, then one cannot say that $b \le \underline{u}$, or $w_V \le w_{M_1^U}$ in their notation.) \end{remark}

Finally we provide a
terminating case of the suggested recursion, corresponding to \citet[Cor. 3.2]{spouge2003} for the least squares case. 
However, our generalization does not come as a simple corollary to \Cref{thm:3_2}, because of the possible nonuniqueness of the solution.
\begin{theorem}\label{thm:constantfit}
	Suppose $b \in \Cf_g(X)$.
	If the empty set is a maximal upper set and a minimal lower set with respect to $b$ at the same time, then 
$b\mathbf{1} \in \argmin_{\hat{g}\in\mathcal{I}_X} \mathcal{J}_X(g, \hat{g})$ solves problem \eqref{eqn:isotonic}.
\end{theorem}
\begin{proof}
	First note that $b\in\Cf_g(X)$ means $0 \in \Sigma_b(X) = -\sum_{x\in X}\partial\ell(g(x), b)$. We want to show
	\begin{equation}\label{eqn:optimality}
		\mathcal{J}_X(g, b\mathbf{1}) \le \mathcal{J}_X(g, \hat{g}), \quad \forall \hat{g} \in \mathcal{I}_X
		.
	\end{equation}

	Since $\emptyset$ is maximal, we see that 
	\begin{equation}\label{eqn:maximal}
		\min\Sigma_b(U) = -\max\sum\nolimits_{x\in U}\partial\ell(g(x), b) \le 0 = \Sigma_b(\emptyset)
	\end{equation}
	for any upper set $U\in\mathcal{U}_X$.
	Thus $f_{U}(y) \triangleq \sum_{x\in U}\ell(g(x), y)$ has a nonnegative maximum subgradient at $y=b$. This means that the convex function $f_U$ is monotone nondecreasing or possesses its minimum at $y=b$. 
In either case,
	\begin{equation}\label{eqn:feasible_U}
		\sum\nolimits_{x\in U}\!\ell(g(x),\! b_1) \!=\! f_U(b_1)
		\!\le\!
		f_U(b_2) \!=\! \sum\nolimits_{x\in U}\!\ell(g(x),\! b_2)
	\end{equation}
	for $b \le b_1 \le b_2$.
	Now since $\emptyset$ is also a minimal lower set, $$
		\max\Sigma_b(U) = -\min\sum\nolimits_{x\in L}\partial\ell(g(x), b) \ge 0
	$$
	for any lower set $L \in \mathcal{L}_X$.
	Thus
the convex function $f_L(y) \triangleq \sum_{x\in L}\ell(g(x), y)$ is monotone nonincreasing or possesses its minimum at $y=b$. 
In either case,
	for $b_2 \le b_1 \le b$,
	\begin{equation}\label{eqn:feasible_L}
		\sum\nolimits_{x\in L}\!\ell(g(x),\! b_1) \!=\! f_L(b_1)
		\!\le\!
		f_L(b_2) \!=\! \sum\nolimits_{x\in L}\!\ell(g(x),\! b_2)
		.
	\end{equation}

	For a given $\hat{g}\in\mathcal{I}_X$, construct a strictly decreasing sequence $\{\underline{b}_k\}$ such that $\underline{b}_0 = b$ and $\underline{b}_{k+1} = \max_{x\in[\hat{g} < \underline{b}_k]}\hat{g}(x)$. Since $X$ is finite, $\{\underline{b}_k\}$ is a finite sequence. Let the length of this sequence be $l$.
	Analogously construct a finite, strictly increasing sequence $\{\bar{b}_k\}$ of length $m$: $\bar{b}_0 = b$ and $\bar{b}_{k+1} = \min_{x\in[\hat{g} > \bar{b}_k]}\hat{g}(x)$.
	Then we can expand $\hat{g}$ as
	\begin{align*}
		\hat{g} &= b\chi_{[\hat{g}\le b]\setminus[\hat{g}\le\underline{b}_1]}
			+ \sum\nolimits_{i=1}^{l-1}\underline{b}_i\chi_{[\hat{g}\le\underline{b}_i]\setminus[\hat{g}\le\underline{b}_{i+1}]}
			+ \underline{b}_l\chi_{[\hat{g}\le\underline{b}_l]}
			\\
		&\quad\quad
			+ \sum\nolimits_{i=1}^{m-1}\bar{b}_i\chi_{[\hat{g}>\bar{b}_{i-1}]\setminus[\hat{g}>\bar{b}_{i}]}
			+ \bar{b}_m\chi_{[\hat{g}>\bar{b}_{m-1}]}
		,
	\end{align*}
	which entails the following along with $[\hat{g}\le\underline{b}_i] \supset [\hat{g}\le\underline{b}_{i+1}]$ and $[\hat{g} > \bar{b}_{i-1}] \supset [\hat{g} > \bar{b}_i]$.
	\begin{align*}
		\mathcal{J}_X(g, \hat{g}) 
		&= 
		\sum\nolimits_{x\in X}\ell(g(x), \hat{g}(x))
		\\
		&= 
		\sum\nolimits_{x\in[\hat{g}\le b]}\ell(g(x), b)
		\\
		&~ + \sum_{i=1}^{l}\big[
			-\!\!\!\!\sum_{x\in[\hat{g}\le\underline{b}_{i}]}\ell(g(x), \underline{b}_{i\!-\!1})
		+\!\!\sum_{x\in[\hat{g}\le\underline{b}_i]}\ell(g(x), \underline{b}_{i}) \big]
		\\
		&~
		+ \sum\nolimits_{x\in[\hat{g}>b]}\ell(g(x), \bar{b}_1)
		\\
		&~ + \sum_{i=2}^{m}\big[
			-\!\!\!\!\sum_{x\in[\hat{g}>\bar{b}_{i\!-\!1}]}\!\!\!\ell(g(x), \bar{b}_{i\!-\!1})
		+\!\!\sum_{x\in[\hat{g}>\bar{b}_{i\!-\!1}]}\!\!\!\ell(g(x), \bar{b}_{i}) \big]
		\\
		&\ge
		\sum\nolimits_{x\in[\hat{g}\le b]}\ell(g(x), b)
		+
		\sum\nolimits_{x\in[\hat{g}>b]}\ell(g(x), \bar{b}_1)
		\\
		&\ge
		\sum\nolimits_{x\in[\hat{g}\le b]}\ell(g(x), b)
		+
		\sum\nolimits_{x\in[\hat{g}>b]}\ell(g(x), b)
		\\
		&=
		\sum\nolimits_{x\in X}\ell(g(x), b)
		= \mathcal{J}_X(g, b\mathbf{1})
		.
	\end{align*}
	The first inequality follows from
	inequality \eqref{eqn:feasible_U}:
	$$
		\sum\nolimits_{x\in[\hat{g}>\bar{b}_{i}]}\ell(g(x), \bar{b}_{i+1})
		\ge
		\sum\nolimits_{x\in[\hat{g}>\bar{b}_{i}]}\ell(g(x), \bar{b}_{i})
	$$
	since $[\hat{g}>\underline{b}_i]$ is an upper set and
	$\bar{b}_{i+1} > \bar{b}_{i} \ge b$,
	and from 
	inequality \eqref{eqn:feasible_L}:
	$$
		\sum\nolimits_{x\in[\hat{g}\le\underline{b}_i]}\ell(g(x), \underline{b}_{i})
		\ge
		\sum\nolimits_{x\in[\hat{g}\le\underline{b}_i]}\ell(g(x), \underline{b}_{i-1})
	$$
	since $[\hat{g}<\underline{b}_i]$ is a lower set and
	$\underline{b}_i < \underline{b}_{i-1} \le b$. 
	The second inequality is also due to inequality \eqref{eqn:feasible_U}.
Thus inequality \eqref{eqn:optimality} holds and $b\mathbf{1}$ solves problem \eqref{eqn:isotonic}.
\end{proof}
\begin{corollary}\label{cor:constantfit}
	If $X$ is a maximal upper set and the empty set is a minimal lower set, or conversely, $X$ is a minimal lower set and the empty set is a maximal upper set, then the conclusion of \Cref{thm:constantfit} follows.
\end{corollary}
\begin{proof}
	If $X$ is a maximal upper set, then $\min\Sigma_b(X) \ge \Sigma_b(\emptyset)=0$. Since $0 \in \Sigma_b(X)$, we know that $\min\Sigma_b(X) \le 0$. Therefore $\min\Sigma(b) = 0 = \Sigma_b(\emptyset)$ and $\emptyset$ is also a maximal upper set. 
	A dual argument reveals that if $X$ is a minimal lower set then so is $\emptyset$.
\end{proof}

 \section{Modified GIRP algorithm}\label{sec:girp}
\subsection{Determining the constant fit}
\textcolor{red}{The recursive binary partitioning procedure suggested by \Cref{thm:decomp,thm:3_2,thm:constantfit} (see the beginning of \Cref{sec:decomposition}) is not fully specified since it depends on the value of the constant fit $b\in\Cf_g(S)$ for each partition $S$ and 
there may be infinitely many choices for $b$.
Which value among $\Cf_g(S)$ should be fitted is unclear.
The chosen constant fits determine $g^*\in\mathbb{R}^X$ as a potential solution to problem \eqref{eqn:isotonic}. However, it is not guaranteed that $g^\ast$ observes the isotonic constraints, i.e., $g^\ast\in\mathcal{I}_X$.
This is because \Cref{thm:3_2} is not constructive and only guarantees the \emph{existence} of a function whose restrictions to a complementary pair qualify that the latter is a projection pair. 
Therefore an additional step of determining a valid constant fit for each partition is necessary;
a careless choice may result in undesirable outcomes as seen in \Cref{ex:counter}.}

To achieve this aim,
consider the following recursive partitioning strategy.
\begin{quote}
	\textbf{(RP)}	In the beginning of the recursive partitioning procedure, fix $b \in \Cf(X)$. For a nontrivial complementary pair $(L, U)$ in which $U$ is a maximal upper set, choose $b_L \in \Cf(L)$ and $b_U \in \Cf(U)$ \emph{closest} to $b$. Repeat this procedure recursively until 
the terminating case occurs.
\end{quote}
Once $b \in \Cf(X)$ is determined, then the candidate solution $g^{\ast}$ to problem \eqref{eqn:isotonic} is unambiguously defined solely based on $X$.
The following result states that the $g^{\ast}$ so selected fulfills the projection pair requirement, and is indeed optimal.

\begin{theorem}\label{prop:correctness}
	The resulting $g^{\ast}$ obtained by applying procedure \textbf{(RP)} for a given $b \in \Cf(X)$ is a solution to problem \eqref{eqn:isotonic}.
	Furthermore, if $S \subset X$ is the set corresponding to a node in the binary tree induced by the procedure, then $g^{\ast}|_S$ minimizes $\mathcal{J}_S(g, \cdot)$ on $\mathcal{I}_S$.
	If this node is non-leaf and $(L, U)$ is the complementary pair partitioning $S$, then for $b_S$ that is the constant fit assigned to this node, we have
	$g^{\ast}|_L \le b_S\mathbf{1} \le g^{\ast}|_U$.
\end{theorem}
\begin{proof}
Procedure \textbf{(RP)} induces a binary tree.
Let us apply induction on the height $k$ of the subtree rooted at the node representing $S$.

For $k=1$, the node representing $S$ is a leaf node, 
hence $g^{\ast}|_S = b_S\mathbf{1}$ 
and the empty set is both maximal upper set and minimal lower set. 
Thus, by \Cref{thm:constantfit}, $g^{\ast}|_S$ minimizes $\mathcal{J}_S(g, \cdot)$ on $\mathcal{I}_S$.
For $k=2$, let $b_L$ and $b_U$ be constant fits for $L$ and $U$, respectively. The nodes representing these sets are both leaves. Thus $g^{\ast}|_U = b_U\mathbf{1}$ minimizes $\mathcal{J}_U(g,\cdot)$. If $b_U < b_S$, then since $\max(b_S, b_U) = b_S$, from the proof of \Cref{thm:3_2}, we see $b_S\mathbf{1}$ also minimizes $\mathcal{J}_U(g,\cdot)$. The difference between the constant fits for the nodes representation $S$ and $U$ is zero for the latter minimizer but is positive for $g^{\ast}|_U$. This contradicts that $g^{\ast}|_U$ is obtained by \textbf{(RP)}, hence $b_S \le b_U$. A similar arguments shows that $b_L \le b_S$. In other words $g^{\ast}|_L \le b_S\mathbf{1} \le g^{\ast}|_U$ and by \Cref{thm:decomp}, $g^{\ast}|_S = g^{\ast}|_L \cup g^{\ast}|_U$ minimizes $\mathcal{J}_S(g, \cdot)$.

Assume that the conclusion of the theorem holds for $k \le h$.
If the height of the subtree rooted at the node representing $S$ is $h+1$, then its left and right subtrees, rooted at the nodes representing $L$ and $U$ respectively, have heights at most $h$. Then, by the induction hypothesis, $g^{\ast}|_L$ and $g^{\ast}|_U$ minimize $\mathcal{J}_L(g, \cdot)$ on $\mathcal{I}_L$ and $\mathcal{J}_U(g, \cdot)$ on $\mathcal{I}_U$ respectively.
Let $\underline{u} = \min_{x\in U}g^{\ast}|_U(x)$. 
Observe that $\underline{u}$ is the constant fit for the leftmost leaf node of the right subtree. If $b_U$ is the constant fit for the root of this subtree, then again by the induction hypothesis, $\underline{u} \le b_U$,
and
along the path from the leftmost node to the root of the right subtree, the constant fits increase from $\underline{u}$ to $b_U$.

We want to show that $b_S\mathbf{1} \le g^{\ast}|_U$, or $b_S \le \underline{u}$.
If to the contrary $\underline{u} < b_S$, then replace all the constant fits for the nodes on the right subtree that are less than $b_S$ with $b_S$. The resulting leaf nodes then defines a new function $\hat{g}_U = \max(b_S\mathbf{1}, g^{\ast}|_U)$.  From the proof of \Cref{thm:3_2}, we see $\hat{g}_U$ also minimizes $\mathcal{J}_U(g, \cdot)$ on $\mathcal{I}_U$.
If $b_U < b_S$, then the constant fits for the nodes along the path from the leftmost node to that representing $U$ in the modified tree are all $b_S$. The difference between the constant fits for the nodes representing $S$ and $U$ are zero in the tree for $\hat{g}_U$, whereas this value is positive for $g^{\ast}|_U$, 
which is a contradiction (see the $k=2$ case).
If $\underline{u} < b_S \le b_U$, then the constant fit for the leftmost node is replaced by $b_S$. The difference between the constant fits for this node and that representing $U$ decreases from $b_U - \underline{u}$ for $g^{\ast}|_U$ to $b_U - b_S$ for $\hat{g}_U$. Thus along the path there is a parent-left child pair whose difference of the constants fits strictly  decreases. This also contradicts that $g^{\ast}|_U$ is obtained by \textbf{(RP)}.

A similar argument yields $g^{\ast}|_L \le b_S\mathbf{1}$. Then, by \Cref{thm:decomp}, $g^{\ast}|_S = g^{\ast}|_L \cup g^{\ast}|_U$ minimizes $\mathcal{J}_S(g, \cdot)$.
\end{proof}

\subsection{Computing a maximal upper set}
\textcolor{red}{The remaining task is to compute a maximal upper set (or a minimal lower set).
In the least squares case it is known that this problem is converted to a network flow problem \citep{picard1976maximal,spouge2003,luss2012}. 
Since our notion of maximal upper set is more general (\Cref{def:maximalupper}), we need a different procedure.}

Chasing the definition we may solve the following optimization problem
in some fashion: 
\begin{equation}\label{eqn:maximalupperset}
	\max\nolimits_{U\in\mathcal{U}_X}\min\Sigma_b(U)
	.
\end{equation}
However, if the optimal value of problem \eqref{eqn:maximalupperset} is zero, 
then although this certifies that $\emptyset$ is a maximal upper set, it is unclear if $\emptyset$ is a minimal lower set as well, required by \Cref{thm:constantfit} to terminate the recursion. Hence in this case we also need to solve the minimal lower set problem
\begin{equation}\label{eqn:minimallowerset}
	\min\nolimits_{L\in\mathcal{L}_X}\max\Sigma_b(L)
	.
\end{equation}
If the optimal value is zero, then this is the certificate that $\emptyset$ is minimal and we can terminate the procedure. Otherwise, the optimal value is strictly negative, hence the minimizing lower set $L$ is nonempty (and minimal).

As an alternative to the above two-step procedure, we consider a one-step procedure of solving
\begin{equation}\label{eqn:painsky2}
	\max_{U\in\mathcal{U}_X, L\in\mathcal{L}_X, L\cap U=\emptyset}\{\min\Sigma_b(U) - \max\Sigma_b(L)\} 
	.
\end{equation}
Note that the lower set $L$ is a subset of $X\setminus U$. Hence it is not fully determined by $U$.
We can show that this optimization problem finds a pair of maximal upper set and minimal lower set simultaneously.
\begin{proposition}\label{prop:correctness2}
	Let $(L^{\star}, U^{\star})$, where $L^{\star} \in \mathcal{L}_X$ and $U^{\star} \in \mathcal{U}_X$, is a solution to problem \eqref{eqn:painsky2}. Then $L^{\star}$ is a minimal lower set and $U^{\star}$ is a maximal upper set in $X$ with respect to $b\in\Cf(X)$.
\end{proposition}
Thus if both $L^{\star}$ and $U^{\star}$ are empty sets or either one is equal to $X$, then we terminate the algorithm. Otherwise, say, $U^{\star}$ is nonemtpy, then we partition $X$ into a complementary pair $(X\setminus U^{\star}, U^{\star})$ and proceed with the recursion.

In order to prove \Cref{prop:correctness2}, we need the following result, which is of independent interest.
\begin{lemma}\label{lem:disjoint}
	There exists a maximal upper set $U$ and a minimal lower set $L$ in $X$ with respect to $b\in\Cf(X)$ such that $L\cap U = \emptyset$.
\end{lemma}
\begin{proof}
	A maximal upper set $U$ and a minimal lower set $L$ can be found by solving problems \eqref{eqn:maximalupperset} and \eqref{eqn:minimallowerset}, respectively. If $L\cap U = \emptyset$, then there is nothing to prove. Otherwise, let $M = L \cap U \neq \emptyset$. Also let $L' = X\setminus U \in \mathcal{L}_X$ and $U' = X\setminus L \in \mathcal{U}_X$.
	Then, since $U = U' \cup M$ is maximal,
	\begin{align*}
		\min\Sigma_b(U) &= \min\Sigma_b(U') + \min\Sigma_b(M)
		\\
		&\le \min\Sigma_b(U) + \min\Sigma_b(M)
		.
	\end{align*}
	Therefore $\min\Sigma_b(M) \ge 0$. Then, from $L = L' \cup M$,
	\begin{align*}
		\max\Sigma_b(L) &= \max\Sigma_b(L') + \max\Sigma_b(M)
\ge \max\Sigma_b(L')
		,
	\end{align*}
	which means that $L'$ is also a minimal lower set.
	It is clear that $L'\cap U = \emptyset$.
\end{proof}
\begin{proof}[Proof of \Cref{prop:correctness2}]
	Observe that
	\begin{align*}
	\max_{U\in\mathcal{U}_X, L\in\mathcal{L}_X, L\cap U=\emptyset}&\{\min\Sigma_b(U) - \max\Sigma_b(L)\} 
	\\
	\le
	\max\nolimits_{U\in\mathcal{U}_X} \min&\Sigma_b(U) - \min\nolimits_{L\in\mathcal{L}_X}\max\Sigma_b(L)
	.
	\end{align*}
	If $(L^{\star}, U^{\star})$ is a pair of disjoint minimal lower and maximal upper sets, which exist due to \Cref{lem:disjoint}, then this pair is feasible for the optimization problem of the left-hand side of the above inequality.
	Also note that $\min\Sigma_b(U^{\star}) = 
	\max_{U\in\mathcal{U}_X} \min\Sigma_b(U)$ and  $\max\Sigma_b(L^{\star}) = \min_{L\in\mathcal{L}_X}\max\Sigma_b(L)$ by construction. Hence the right-hand side is attained as the maximum of problem \eqref{eqn:painsky2}.
	If a pair $(L^{\dagger}, U^{\dagger})$ attains the maximum but $L^{\dagger}$ is not a minimal lower set or $U^{\dagger}$ is not a maximal upper set, then 
	\begin{align*}
		\min\Sigma_b(U^{\star}) - \max\Sigma_b(L^{\star})
		&=
		\min\Sigma_b(U^{\dagger}) - \max\Sigma_b(L^{\dagger})
		\\
		&< 
		\min\Sigma_b(U^{\star}) - \max\Sigma_b(L^{\star})
		,
	\end{align*}
a contradiction.
Thus any solution to problem \eqref{eqn:painsky2} is a pair of a minimal lower set and a maximal upper set.
\end{proof}

Problem \eqref{eqn:painsky2} is equivalent to the partitioning rule \eqref{eqn:painsky} of \citep[eqn. 11]{painsky2016}, which explicitly involves an additional set $M^{\star}=X\setminus L^{\star}\setminus U^{\star}$. Thus we have shown here that the partitioning rule of \citet{painsky2016} in fact finds a pair of minimal lower and maximal upper sets simultaneously. 
While \citet{painsky2016} suggests partitioning $X$ into three sets $(L^{\star}, M^{\star}, U^{\star})$, \Cref{thm:3_2} asserts that binary partitioning into either $(L^{\star}, X\setminus L^{\star})$ or $(X\setminus U^{\star}, U^{\star})$ suffices.
How to carry out the optimization by formulating problem \eqref{eqn:painsky2} as a linear programming (LP) is discussed in detail in \citep[eqn. (12) and \S2]{painsky2016}.
In particular, it converts to a maximum flow problem for which many efficient algorithm exist \citep{boykov2004experimental}. 

If the coordinate loss functions are differentiable, then by \Cref{rem:spouge}, it suffices to solve problem \eqref{eqn:maximalupperset}, which allows a simpler LP formulation \citep{picard1976maximal}.
In fact problem \eqref{eqn:painsky2} reduces to problem \eqref{eqn:maximalupperset} in this case.

\subsection{Algorithm}
\Cref{alg:rp} presents a pseudo-code description of \textbf{(RP)}.
\Cref{prop:correctness} guarantees that any pruning of the binary tree induced by this algorithm yields a sub-optimal solution to \eqref{eqn:isotonic} observing the isotonic constraints. If $\Cf_g(S)$ is a singleton for each input $S$, then the original GIRP \citep{luss2014} 
is recovered.
\begin{algorithm}[h!]\small
\caption{Modified GIRP}\label{alg:rp}
  \begin{algorithmic}
\Function{mGIRP}{$g$, $S$, $b_P$}
		\If{$b_P$ is defined}
			\State $b \gets \argmin\{|b_P - v|: v \in \Cf(S)\}$
		\Else
			\State Pick $b \in \Cf_g(S)$
		\EndIf
\State Solve problem \eqref{eqn:painsky2} to find a complementary pair $(L, U)$
		\If{$\min\Sigma_b(U)==0$ and $\max\Sigma_b(L)==0$}
			\State \Return{$g^{\star}|_S = b\mathbf{1}$}
		\EndIf
		\State $g^{\star}|_L \gets$ mGIRP$(g, L, b)$ 
		\State $g^{\star}|_U \gets$ mGIRP$(g, U, b)$
		\State \Return{$g^{\star}|_S = g^{\star}|_L \cup g^{\star}|_U$}
   \EndFunction
  \end{algorithmic}
\end{algorithm}

 \section{Illustrative examples}\label{sec:examples}
\begin{example}[Hinge loss]\label{ex:hinge}
	Let $X=\{1, 2, 3\}$, $g(1)=g(2)=1$, and $g(3)=-1$. 
	The isotonicity constraint is the total order $\hat{g}(1) \ge \hat{g}(2) \ge \hat{g}(3)$.
	If we employ the hinge loss
	$\ell(g(x), y)=\max(0, 1 - y g(x))$, then problem \eqref{eqn:isotonic} has a trivial solution of $\hat{g} = g$, with the optimal value of zero.
	In fact it has infinitely many solutions of the form  $\hat{g}(1)=\hat{g}(2)=\alpha$ and $\hat{g}(3)=\beta$ where $\alpha \ge 1$ and $\beta \le -1$.
	To see if \Cref{alg:rp} gives a correct solution,
	observe that $\Cf_g(X) = \argmin_{y\in\mathbb{R}}\{2\max(0, 1 - y) + \max(0, 1 + y)\}= 1 =: b$.
	Then,
$\partial\ell(g(1), b) = \partial\ell(g(2), b) = [-1, 0]$,
$\partial\ell(g(3), b) = 1$,
$\mathcal{U}_X = \{X, \{1, 2\}, \{1\}, \emptyset\}$,
	and $\mathcal{L}_X = \{\emptyset, \{3\}, \{2, 3\}, X\}$.
	In order to solve problem \eqref{eqn:painsky2}, let us tabulate $\Sigma_b(U)$ and $\Sigma_b(U)$ in \Cref{tbl:hinge} for $U \in\mathcal{U}_X$ and $L\in\mathcal{L}_X$.
	The optimal value of problem \eqref{eqn:painsky2} is $1$, occurring at $(L, U)=(\{3\}, \{1, 2\}), (\{3\}, \{1\}), (\{3\}, \emptyset)$. Verify that each $L$ ($U$) is minimal (maximal). 
	Since $\max\Sigma_b(L)=-1$, \Cref{alg:rp} proceeds with partitioning $X$. The possible partitions are $(L, U) = (\{3\}, \{1, 2\})$ and $(\{2, 3\}, \{1\})$.

	With the first partition, $L=\{3\}$ needs no further partitioning. 
	Since $\Cf_g(L) = (-\infty, -1]$, \Cref{alg:rp} finds $g^{\star}|_L = (-1)\mathbf{1}$.
To test $U=\{1, 2\}$,
	solve problem \eqref{eqn:painsky2} with $X$ replaced by $U$. 
	Then we see
	$\Cf_g(U) = \argmin_{y\in\mathbb{R}}\{2\max(0, 1 - y)\} = [1, \infty)$ and \Cref{alg:rp} chooses $b=1$.
	We tabulate $\Sigma_b(U')$ and $\Sigma_b(L')$ for $U'\in\mathcal{U}_U = \{U, \{1\}, \emptyset\}$ and $L'\in\mathcal{L}_U=\{\emptyset, \{2\}, U\}$ in \Cref{tbl:hinge}.
	The optimal value of problem \eqref{eqn:painsky2} is $0$, attained at $(L, U)=(\emptyset, \{1, 2\}), (\emptyset, \{1\}), (\emptyset, \emptyset)$. For each pair $\max\Sigma_b(L')=0=\min\Sigma_b(U')$, so \Cref{alg:rp} terminates, yielding $g^{\star}|_U = (1)\mathbf{1}$ and thus $g^{\star} = g$.

	With the second partition, for $U=\{1\}$ we have $g^{\star}|_U = (1)\mathbf{1}$. 
	We then see $\Cf_g(L) = [-1, 1]$ for $L=\{2, 3\}$, and \Cref{alg:rp} assigns $b=1$ to $L$.
	For $U'\in  \mathcal{U}_L = \{\emptyset, \{2\}, \{2, 3\}\}$ and $L'\in\mathcal{L}_L = \{\emptyset, \{3\}, \{2, 3\}\}$, we build \Cref{tbl:hinge}.
	We see that the optimal value of problem \eqref{eqn:painsky2} with $X$ replaced by $L$ is $1$, occurring at $(L', U') = (\{3\}, \{2\}), (\emptyset, \{3\})$. Since $\max\Sigma_b(L')=-1$, we further partition $L$ into $(L', U') = (\{3\}, \{2\})$.
	It is obvious that $-1$ should be fitted to $L'$ and $1$ to $U'$, yielding $g^{\star}|_L(2) = 1$ and $g^{\star}|_L(3) = -1$.
	Again we have $g^{\star} = g$.
	Note even if the final split is not made, the resulting function $\hat{g} = (1)\mathbf{1}$ is isotonic.
\end{example}
\begin{table}\caption{Table for computing partitions in \Cref{ex:hinge}. ``up''=upper set, ``low''=lower set.}\label{tbl:hinge}
	\scriptsize
\begin{center}
	\begin{tabular}{cccccc}
	\toprule
set & up & low & $\Sigma_b(\text{up})$ & $\Sigma_b(\text{low})$ & $\Sigma_b(\text{up}) \!-\! \Sigma_b(\text{low})$ \\
	\midrule
	\multirow[c]{10}{*}{$X$} &
	\{1, 2, 3\}    & $\emptyset$ & [-1, 1] & 0       & [-1, 1] \\
	& \{1, 2\}	   & $\emptyset$ & [0, 2]  & 0       & [0, 2]  \\
	& \{1, 2\}	   & \{3\}       & [0, 2]  & -1      & [1, 3]  \\
	& \{1\}	       & $\emptyset$ & [0, 1]  & 0       & [0, 1]  \\
	& \{1\}	       & \{3\}       & [0, 1]  & -1      & [1, 2]  \\
	& \{1\}	       & \{2, 3\}    & [0, 1]  & [-1, 0] & [0, 2]  \\
	& $\emptyset$	   & $\emptyset$ & 0       & 0       & 0       \\
	& $\emptyset$	   & \{3\}       & 0       & -1      & 1       \\
	& $\emptyset$	   & \{2, 3\}    & 0       & [-1, 0] & [0, 1]  \\
	& $\emptyset$	   & \{1, 2, 3\} & 0       & [-1, 1] & [-1, 1] \\
	\midrule
\multirow[c]{6}{*}{$U$} &
	\{1, 2\}     & $\emptyset$ & [0, 2]  & 0       & [0, 2]  \\
	& \{1\}	     & $\emptyset$ & [0, 1]  & 0       & [0, 1]  \\
	& \{1\}	     & \{2\}       & [0, 1]  & [0, 1]  & [-1, 1] \\
	& $\emptyset$	 & $\emptyset$ & 0       & 0       & 0       \\
	& $\emptyset$	 & \{2\}       & 0       & [0, 1]  & [-1, 0] \\
	& $\emptyset$	 & \{1, 2\}    & 0       & [0, 2]  & [-2, 0] \\
	\midrule
\multirow[c]{6}{*}{$L$} &
	\{2, 3\}     & $\emptyset$ & [-1, 0] & 0       & [-1, 0] \\
	& \{2\}	     & $\emptyset$ & [0, 1]  & 0       & [0, 1]  \\
	& \{2\}	     & \{3\}       & [0, 1]  & -1      & [1, 2]  \\
	& $\emptyset$	 & $\emptyset$ & 0       & 0       & 0       \\
	& $\emptyset$	 & \{3\}       & 0       & -1      & [-1, 0] \\
	& $\emptyset$	 & \{2, 3\}    & 0       & [-1, 0] & [0, 1]  \\
	\bottomrule
	\end{tabular}
\end{center}
\end{table}

\begin{example}[\Cref{ex:counter} revisited]
	\color{red}
	Recall that the Huber loss with $\delta = 0.9$ is used in \Cref{ex:counter}. Then $\Cf_g(X) = \argmin_{y\in\mathbb{R}}\sum_{x\in X}\ell_{\text{H},\delta}(g(x), y)=4.0:=b$ after a tedious but straightforward calculation.
	We the solve problem \eqref{eqn:painsky2} for this value of $b$ by solving the LP relaxation (which is tight \citep{luss2014}):
$$
	\begin{array}{ll}
	\text{minimize} & \sum_{x\in X}(-\frac{\partial\ell(g(x), y)}{\partial y}|_{y=b}) z_x  \\
\text{subject to} & 0 \le z_x \le 1, \quad x \in X ~~\text{and}
	\quad
	z_x \le z_y, \quad x \preceq y 
	.
	\end{array}
$$
	using the COSMO solver \citep{garstka2019}.
	This yields the pair $(L, U)$ as given in \Cref{ex:counter}, with $\Sigma_b(L)=0.9=-\Sigma_b(U)$.

	For the $U$ we have $\Cf_g(U) = 4.1$. 
	Solving problem \eqref{eqn:painsky2} with $X$ replaced by $U$ using COSMO partitions the latter into the pair $(L', U')$ in \Cref{ex:counter}, 
with $\Sigma_{4.1}(L')=0.9=-\Sigma_{4.1}(U')$. 
	To $L'$ we fit $\Cf_g(L') = 4.0$ and to $U'$ we fit $\Cf_g(U') = 5.0$. 
	Further solving problem \eqref{eqn:painsky2} leads to an empty maximal upper set for both sets, hence there is no more split.

Similarly we have $\Cf_g(L) = 3.78$. 
	COSMO partitions $L$ into the pair $(L'', U'')$ given in \Cref{ex:counter}, 
with $\Sigma_{3.78}(L'')=0.12=-\Sigma_{3.78}(U'')$.
	On $L''$ we have $\Cf_g(L'') = 3.75$. Further solving problem \eqref{eqn:painsky2} leads to an empty maximal upper set, hence $L''$ needs no more split.
	On $U''$ we see that $\Cf_g(U'') = [3.9, 4.1]$ and \Cref{alg:rp} selects $b=3.9$. 
	Solving problem \eqref{eqn:painsky2} results in $(L''', U''')$ in \Cref{ex:counter}, with $\Sigma_{3.9}(L''') = 0 = -\Sigma_{3.9}(U''')$.
	This means that $U''$ itself is maximal. 
	From \Cref{cor:constantfit} partitioning of $U''$ is not necessary. This is equivalent to fitting $3.9$ to $U''$, $L'''$, and $U'''$ in \Cref{fig:counter}.

	The resulting fit $\hat{g}$ satisfies  the isotonic constraints. 
\end{example}

 \section{Conclusion}\label{sec:conclusion}
In this paper we have focused on the possible nonuniqueness of the solution to the isotonic regression problem \eqref{eqn:isotonic}, which arises in many important coordinate loss functions, in case the solution exists. The lack of uniqueness of the solution raises a question on the correctness of the generalized isotonic recursive partitioning (GIRP) algorithm \citep{luss2014,painsky2016}, which is a greedy algorithm that solves a simple graph cut problem 
in every step to partition the observations into several groups, while maintaining the isotonicity constraint.
Our result, which is a nontrivial extension of the least-squares case \citep{spouge2003}, however shows that there is at least one solution to \eqref{eqn:isotonic} that can be found by a minimally modified GIRP algorithm,
where the modification includes partitioning always into two sets (instead of three).
Hence we have genuinely shown the correctness of the GIRP algorithm.
The issue of existence and uniqueness of the solution has been obscured in the literature probably because the work \citep{painsky2016} focuses on the logistic
loss function with the lasso penalty, which is nondifferentiable but nonetheless admits a unique solution.
Our analysis demonstrates that a care must be taken to deal with nonunique solutions.

\singlespacing

\bibliographystyle{chicago}
\bibliography{girp.bib}

\end{document}